\documentclass{article}

     \PassOptionsToPackage{numbers, compress}{natbib}
\usepackage[preprint]{neurips_2020}

\usepackage[utf8]{inputenc} % allow utf-8 input
\usepackage[T1]{fontenc}    % use 8-bit T1 fonts
\usepackage{hyperref}       % hyperlinks
\usepackage{url}            % simple URL typesetting
\usepackage{booktabs}       % professional-quality tables
\usepackage{amsfonts}       % blackboard math symbols
\usepackage{nicefrac}       % compact symbols for 1/2, etc.
\usepackage{microtype}      % microtypography
\usepackage{graphicx}
\usepackage{extpfeil}
\usepackage{subfigure}
\usepackage{amsmath}
\usepackage{makecell}
\usepackage{multirow}
\usepackage{amsfonts}
\usepackage{varwidth}
\usepackage{caption}
\usepackage{amsthm,amsmath,amssymb}
\usepackage{mathrsfs}
\usepackage{bm}
\usepackage{color,xcolor}
\usepackage{booktabs} % for professional tables

\title{Technical Note: Game-Theoretic Interactions of Different Orders}

\author{%
  Hao Zhang\footnotemark[1]\thanks{Contribute equally to this paper.} \\
  Shanghai Jiao Tong University\\
  \texttt{1603023-zh@sjtu.edu.cn} \\
 \And
  Xu Cheng\footnotemark[1] \\
 Shanghai Jiao Tong University\\
\texttt{xcheng8@sjtu.edu.cn} \\
\And
   Yiting Chen\footnotemark[1]\\
 Shanghai Jiao Tong University\\
\texttt{sjtucyt@sjtu.edu.cn} \\
 \And
  Quanshi Zhang\thanks{Correspondence.} \\
 Shanghai Jiao Tong University\\
\texttt{zqs1022@sjtu.edu.cn} \\
}

\begin{document}

\maketitle
\section{Introduction}
Deep neural networks (DNNs) have exhibited significant success in various tasks, and the interpretability of DNNs has received increasing attention in recent years.
Quantifying the interaction between input variables of the DNN provides a new perspective to explain the signal processing encoded in a DNN~\cite{bien2013a, sorokina2008detecting, murdoch2018beyond, singh2018hierarchical, jin2019towards, janizek2020explaining, sundararajan2017axiomatic, tsang2018detecting, lundberg2018consistent, Grabisch1999AnAA, dhamdhere2019shapley, zhang2020interpreting, zhang2020hierarchical, wang2020unified, Tsang2020Feature, cuilearning, zhang2020dropout}.

Given a DNN, we quantify interactions of different orders between two input variables of the DNN during the inference process.
% Each input variable of a DNN usually does not work individually. Instead, input variables may cooperate with other input variables to make inferences.
% Previous studies mainly focused on the interaction between two variables~\cite{lundberg2018consistent, singh2018hierarchical, murdoch2018beyond, janizek2020explaining}.
Previous studies have been proposed to measure the interaction based on game theory~\cite{Grabisch1999AnAA, dhamdhere2019shapley}.
\citet{Grabisch1999AnAA} proposed the interaction index to measure $2^n$ different interaction values for all $2^n$ combinations of $n$ variables.
\citet{dhamdhere2019shapley} extended the study of~\cite{Grabisch1999AnAA}, and proposed the Shapley-Taylor interaction index.
Compared with previous studies, we focus on the interaction between two input variables, and disentangle the interaction into interaction components of different orders, which reflect inter-variable effects \emph{w.r.t.} contexts of different scales.
% In comparison to the unaffordable computational cost, our research summarizes all $2^n$ interactions into a single metric, which provides a global view to understand DNNs.

In fact, this is just a technical note for multi-order interactions. In another paper, we have formally introduced details and the proof of multi-order interactions, and we have used the multi-order interaction to explain more signal-processing properties of DNNs.

In this study, we define interaction components of different orders between two input variables based on game theory. Given a DNN with $n$ input variables, each input variable can be viewed as a player. In this way, we use $N=\{1,2,3,\cdots,n\}$ to represent the set of all players (indices of all input variables). Let $v$ denote the output of the DNN, which can be considered as a game function. The function $v$ maps a set of players to a scalar value, \emph{i.e.} $v:2^N\rightarrow \mathbb{R}$, where $2^N$ denotes all possible subsets of players in the set $N$.
Let $\phi(i)$ denote the contribution of the player $i$ to the game, which can be computed as the Shapley value of the player $i$. The Shapley value is widely regarded as a unique unbiased estimation of the contribution that satisfies the \emph{linearity, dummy, symmetry}, and \emph{efficiency} properties~\citep{weber1988probabilistic}.

% {\color{red}The interaction between players $i$ and $j$ is defined as the difference between the contribution of the player $i$ with the player $j$ present and the contribution of the player $i$ with the player $j$ absent, which is denoted by $I(i,j)$.}
The overall interaction $I(i,j)$ between the player $i$ and the player $j$ is defined as the change of $\phi(i)$ when the player $j$ is absent \emph{w.r.t.} the case when the player $j$ is present, \emph{i.e.} $I(i,j) = \phi_{\textrm{w/}\,j }(i) - \phi_{\textrm{w/o}\,j}(i)$.
If $I(i,j) >0$, then we consider players $i$ and $j$ have a positive effect. Accordingly, if $I(i,j) <0$, then we consider players $i$ and $j$ have a negative effect.

% {\color{green}We notice that the contribution of the player $i$ would be different when the player $j$ never participates in the game \emph{w.r.t.} the case when the player $j$ always participates in the game.}
% Thus, the interaction {\color{red}$I(i,j)$} is measured as the difference $\phi_{\text{w/ }j}(i)-\phi_{\text{w/o }j}(i)$.
% If a set of $m$ players $S$ always participates in the game together, these players can be regarded to form a coalition. The coalition can obtain a reward, denoted by $\phi_S$. $\phi_S$ is usually different from the sum of rewards when players in the set $S$ participate in the game individually. The additional reward $\phi_S-\sum_{i\in S}\phi_i$ obtained by the coalition can be quantified as the interaction.
% If $\phi_S-\sum_{i\in S}\phi_i>0$, we consider players in the coalition have a positive effect. Whereas, $\phi_S-\sum_{i\in S}\phi_i<0$ indicates an negative/adversarial effect among the set of variables in $S$.

More specifically, we explicitly decompose $I(i,j)$ into $0$-order, $1$-order, $\dots$, $(n-2)$-order interaction components,~\emph{i.e.} $I(i,j) = \frac{1}{n-1}\sum_{m=0}^{n-2} I^{(m)}(i,j)$, in order to measure the scale of inference patterns encoded in the DNN. Here, the order $m$ indicates the scale of the context involved in the computation of interactions between pixels $i$ and $j$. For example, given a human face, we consider players $i$ and $j$ as two eyes on this face. Besides, we regard other $m$ players, which are included on the face as the context.
The interaction between two eyes depends on such a context of the face, which is measured as $I^{(m)}(i,j)$. Without the face context, there is supposed not to contain such interaction. Therefore, the order $m$ reflects the scale of inference patterns that are encoded in the DNN.
% In this way, $I^{(m)}(i,j)$ measures the interaction between two eyes depending on the context of the face.

% However, $\phi_S-\sum_{i\in S}\phi_i$ mainly measures the interaction in a single coalition, whose interaction is either purely positive or purely negative. In real applications, for the set $S$ with $m$ players, players may form at most $2^m-m-1$ different coalitions.
% Some coalitions have positive interaction effects, while others have negative interaction effects. Thus, interactions of different coalitions can counteract each other.

Contributions of this study can be summarized as follows. (1) In this study, we define and quantify interaction components of different orders. (2) We further prove that interaction components of different orders satisfy several desirable properties.

\section{Algorithm}
\subsection{Shapley values}
Given a game with multiple players, each player is supposed to obtain a high reward. Some players may form a coalition to pursue a high reward. Considering the contribution of each player to the coalition is different, each player should be assigned with different rewards. Let $N$ denote the set of all players, and $2^N\overset{\textrm{def}}{=}\{S|S\subseteq N\}$ indicate all potential subsets of $N$. For each subset of players $S$, a game $v:2^{N} \rightarrow \mathbb{R}$ denotes a scalar reward obtained by $S$. The Shapley value $\phi_{v}(i)$ of the player $i$ represents the \textbf{numerical contribution} of this player to the game $v$, which is defined by~\cite{shapley1953value}.
\begin{equation}
\label{eqn:shapleyvalue}
\phi_{v}(i|N)=\sum \nolimits_{S\subseteq N\setminus\{i\}}\frac{(n-s-1)!s!}{n!}\Delta_{i}v(S),\quad\Delta_{i}v(S)\overset{\textrm{def}}{=}v(S\cup\{i\})-v(S),
\end{equation}
where $|S|=s$ and $|N|=n$.~\citet{weber1988probabilistic} have proven that the Shapley value is a unique unbiased method to fairly allocate overall reward to each player with four properties. For simplicity, we use $\phi(i|N)$ by ignoring the superscript of $\phi_{v}(i|N)$ in the following manuscript without causing ambiguity.

$\bullet\;$\emph{{Linearity property}}: If two independent games can be merged into one game, then the Shapley value of the new game also can be merged, \emph{i.e.} $\forall S \in N$, $\phi_{u}(i|N)=\phi_{v}(i|N)+\phi_{w}(i|N)$; $\forall c \in \mathbb{R}$, $\phi_{c \cdot u} (i|N)= c\cdot \phi_{u}(i|N)$.

$\bullet\;$\emph{{Dummy property}}: The dummy player $i$ is defined as a player satisfying $\forall S\subseteq N\setminus\{i\}$, $v(S\cup\{i\})=v(S)+v(\{i\})$, which indicates that the player $i$ has no interactions with other players in $N$, $\phi(i|N)=v(\{i\})- v(\emptyset)$.

$\bullet\;$\emph{{Symmetry property}}: If $\forall S\subseteq N\setminus\{i\}$, $v(S\cup\{i\})=v(S\cup\{j\})$, then $\phi(i|N)=\phi(j|N)$.

$\bullet\;$\emph{{Efficiency property}}: Overall reward can be assigned to all players, $\sum_{i\in N}\phi(i|N)=v(N) - v(\emptyset)$.

\subsection{The interaction index $\mathcal{I}_v(S)$ \cite{Grabisch1999AnAA} and its extension \cite{zhang2020interpreting}}
\citet{Grabisch1999AnAA} firstly proposed the interaction index $\mathcal{I}_v(S)$. Given a game $v$, and a set of all players $N$. For a subset of players $S\subseteq N$, if players in $S$ cooperate to form a coalition for a high reward, then this coalition can be considered as a new singleton, which is represented using brackets, $[S]$. In this way, the game $v$ can be considered to have $(n-s+1)$ players, and one of them is the singleton $[S]$. $\mathcal{I}_v(S)$ quantifies the marginal reward of $S$, which removes all marginal rewards from all potential subsets of $S$.
\begin{equation}
\label{interaction index}
\mathcal{I}_v(S)=\sum_{T\subseteq N\setminus S}\frac{(n-t-s)!t!}{(n-s+1)!}\Delta_{S}v(T), \quad \Delta_{S}v(T) \overset{\textrm{def}}{=}\sum_{L\subseteq S}(-1)^{l-s}v(L\cup T).
\end{equation}
The physical meaning of the interaction index $\mathcal{I}_v(S)$ is well-described by following recursive properties.

$\bullet\;$\emph{{Recursive property 1}}: Let $\mathcal{I}(S|N)$ denote the interaction index that computed with the set of players $N$. Then $\forall S\subseteq N$, $s>1$, $\mathcal{I}(S|N)=\mathcal{I}([S]|N\setminus S \cup \{[S]\})-\sum_{K\subsetneqq S, K\neq \emptyset}\mathcal{I}(K|N\setminus S \cup K)$.

The recursive property 1 indicates that $\mathcal{I}_v(S)$ contains marginal reward of $S$, and removes all marginal reward of subsets of $S$. For example, let $S\!\!=\!\!\{a,b,c\}$, then $\mathcal{I}_v(S)$ contains the marginal reward obtained by $\{a,b,c\}$, and removes marginal rewards obtained by $\{a,b\}$, $\{a,c\}$, $\{b,c\}$, $\{a\}$, $\{b\}$, and $\{c\}$.

$\bullet\;$\emph{{Recursive property 2}}: The interaction index in $S$ is equal to the interaction among players $S\setminus \{i\}$ with the presence of the player $i$ minus the interaction among players $S\setminus \{i\}$ with the absence of the player $i$, where $i$ is an arbitrary player inside $S$. \emph{I.e.} $\forall i\in S$, $\mathcal{I}(S|N)=\mathcal{I}_{\text{w/}\; i}(S\!\setminus\! \{i\}|N\!\setminus\!\{i\})-\mathcal{I}_{\text{w/o}\; i}(S\!\setminus\! \{i\}|N\!\setminus\!\{i\})$, where $\mathcal{I}_{\text{w/}\; i}(S\!\setminus\! \{i\}|N\!\setminus\!\{i\})$ denotes the interaction among $S\!\setminus\! \{i\}$ with the presence of $i$. In this case, we do not consider the player $i$ as an ordinary player that may cooperate with others, but a priori of the game, \emph{i.e.} the player $i$ always participates in the game. $\mathcal{I}_{\text{w/o}\; i}(S\!\setminus\! \{i\}|N\!\setminus\!\{i\})$ denotes the interaction among $S\!\setminus \!\{i\}$ with the absence of $i$. In this case, the player $i$ never participates in the game.

Furthermore,~\citet{Grabisch1999AnAA} have also proven that the interaction index $\mathcal{I}_v(S)$ satisfies following properties. For simplicity, we use $\mathcal{I}(S)$ by ignoring the superscript of $\mathcal{I}_{v}(S)$ in the following manuscript without causing ambiguity.

$\bullet\;$\emph{{Linearity property}}: If $\forall S\subseteq N$, rewards of games $u$, $v$, and $w$ satisfy $u(S)=v(S)+w(S)$, then $\mathcal{I}_u(S)=\mathcal{I}_v(S)+\mathcal{I}_w(S)$; $\forall c \in \mathbb{R}$, $\mathcal{I} _{c \cdot u}(S)= c\cdot \mathcal{I} _{u}(S)$.

$\bullet\;$\emph{{Dummy property}}: If the player $i$ is a dummy player, then $\forall S\subseteq N\setminus\{i\}$, $S\neq \emptyset$, $\mathcal{I}(S\cup\{i\})=0$.

$\bullet\;$\emph{{Symmetry property}}: If $\forall S\subseteq N$ and $i\neq j$, there is $v(S\cup \{i\})=v(S\cup \{j\})$, then $\mathcal{I}(S\cup\{i\})=\mathcal{I}(S\cup\{j\})$.

Moreover, \citet{zhang2020interpreting} proposed $B([S])$ to measure the overall interaction among players in $S$.
\begin{equation}
    B([S])=\phi([S]|N\setminus S\cup\{[S]\})-\sum_{i\in S}\phi(i|N\setminus S\cup \{i\})=\sum_{S'\subseteq S,s'>1}\mathcal{I}(S').
\end{equation}
In this way, the interaction $B([S])$ contains $2^s-s-1$ potential interaction indices inside $S$, where positive and negative interaction indices can counteract each other. Thus, $B'([S])$ is used to reflect the significance of interactions among players inside $S$,
\begin{equation}
    B'([S])=\sum_{S'\subseteq S, s'>1}|\mathcal{I}(S')|=\sum_{S'\subseteq S, s'>1, \mathcal{I}(S')>0}\mathcal{I}(S')-\sum_{S'\subseteq S, s'>1, \mathcal{I}(S')<0}\mathcal{I}(S').
\end{equation}
\citet{zhang2020interpreting} proposed an efficient-yet-approximate method to estimate $B'([S])$. Compared with the interaction index, $B'([S])$ provides a more global view to understand the game.

\subsection{The Shapley-Taylor interaction index $\mathcal{I}^{(k)}_{v}(S)$ \cite{dhamdhere2019shapley}}
\citet{dhamdhere2019shapley} proposed the Shapley-Taylor index, which attributed the model's prediction to interactions among subsets of features up to the size $k$. Specifically, the Shapley-Taylor interaction index is equal to the Taylor Series of the multilinear extension of the set-theoretic behavior of the model. For a fixed ordering $\pi = (i_1,i_2,\dots,i_n)$ and a set of features $S$, the Shapley-Taylor indices $\mathcal{I}^{(k)}_{v}(S|\pi)$ is defined as follows.
\begin{equation}
\label{Shapley-Taylor}
\mathcal{I}^{(k)}_{v}(S|\pi)=\left\{\begin{aligned}
&\Delta_{S}v(\emptyset),\quad\quad  \textrm{if}\; s<k,\\
&\Delta_{S}v(\pi^{S}), \quad\;\textrm{if} \;s=k
\end{aligned}\right.
\end{equation}
The Shapley-Taylor interaction index is computed by averaging all potential ordering $\pi$, as follows.
\begin{equation}
 \mathcal{I}^{(k)}_{v}(S)=\mathbb{E}_{\pi}(\mathcal{I}^{(k)}_{v}(S|\pi))
\end{equation}
\citet{dhamdhere2019shapley} have proven that the Shapley-Taylor interaction index satisfies following properties. For simplicity, we use $\mathcal{I}^{(k)}(S)$ by ignoring the superscript of $\mathcal{I}_{v}^{(k)}(S)$ in the following manuscript without causing ambiguity.

$\bullet\;$\emph{{Linearity property}}: If $\forall S\subseteq N$, three games $u$, $v$, $w$ satisfy $u(S)=v(S)+w(S)$, then $\mathcal{I}^{(k)}_{u}(S)= \mathcal{I}^{(k)}_{v}(S)+ \mathcal{I}^{(k)}_{w}(S)$, and $\mathcal{I}^{(k)}_{c \cdot u}(S)=c\cdot\mathcal{I}^{(k)}_{u}(S)$, where $c$ is a scalar value.

$\bullet\;$\emph{{Dummy property}}: If the player $i$ is a dummy player, then $\forall S\subseteq N$ with $i\in S$, we have $\mathcal{I}^{(k)}(i)=v(\{i\})$, and $\mathcal{I}^{(k)}(S)=0$.

$\bullet\;$\emph{{Symmetry property}}: If $\forall S\subseteq N\setminus\{i,j\}$ and $i\neq j$, there is $v(S\cup \{i\})=v(S\cup\{j\})$, then $\mathcal{I}^{(k)}(i)=\mathcal{I}^{(k)}(j)$.

$\bullet\;$\emph{{Efficiency property}}: The overall reward can be allocated to each potential $S$ with size up to $k$. \emph{I.e.} $\forall v$, $\sum_{S\subseteq N, s\le k}\mathcal{I}^{(k)}(S)=v(N)-v(\emptyset)$.

$\bullet\;$\emph{Interaction Distribution property}:
This property characterizes how interactions are distributed for a class of interaction functions, which model pure interactions and are defined as follows. Given a set $T$, if $T\nsubseteq S$, then the interaction function $v_T(S)=0$; otherwise $v_T(S)=c$, where $c$ is a constant value. In this way, the interaction distribution property is defined as $\forall S\subsetneqq T$, $s<k$, $\mathcal{I}_S^k(v_T)=0$.

\subsection{Interaction of different orders}
In this section, we first extend the original definition of Shapley values into Shapley values of different orders, and introduce interactions between two players based on Shapley values. Then we propose to decompose such interactions into interaction components of different orders.

\textbf{Shapley values of different orders:} Theoretically, $\phi(i|N)$ can be decomposed into Shapley values of different orders, \emph{i.e.} $ \phi(i|N)= \frac{1}{n}\sum_{m=0}^{n-1}\phi^{(m)}(i|N)$. Here, we use $\phi^{(m)}(i|N)$ to measure the numerical contribution of the player $i$ to the game \emph{w.r.t.} the context consisting of $m$ players.
\begin{equation}
\label{eqn:multi_order_shapleyappr}
\phi^{(m)}(i|N)=\mathbb{E}_{S\subseteq N\setminus\{i\},s=m} [ v(S\cup {i})-v(S)],
\end{equation}
where the order $m$ refers to the number of players in $S$. In addition, we have proven that for $m\in \{0,\dots,n-1\}, \phi^{(m)}(i|N)$ has following properties.

$\bullet\;$\emph{Linearity property}: If two games $v$ and $w$ can be combined into a single game, their Shapley values can be added, \emph{i.e.}  $\forall i\in N$, $\phi_u^{(m)}(i|N) = \phi_w^{(m)}(i|N) + \phi_v^{(m)}(i|N)$, and $\phi_{c \cdot u}^{(m)}(S)=c\cdot\phi_{ u}^{(m)}(S)$, where $c$ is a scalar value.

$\bullet\;$\emph{Dummy property}: A player $i\in N$ is considered as a dummy player if $\forall S\subseteq N$, $v(S\cup \{i\}) = v(S)+v(\{i\})$. Thus, the player $i$ has no interactions with other players, \emph{i.e.} $\phi^{(m)}(i|N)=v(\{i\})-v(\emptyset)$.

$\bullet\;$\emph{Symmetry property}: Given two players $i,j\in N$, if these two players have same interactions with all other players $\forall S\subseteq N \backslash \{i,j\}$, $v(S\cup \{i\}) = v(S\cup \{j\})$, then $ \phi^{(m)}(i|N) = \phi^{(m)}(j|N) $.

\textbf{Interactions between two players:}  Given two players $i$ and $j$, if these two players always participate in the game together or always do not participate in the game together, we can roughly consider these two players cooperate with each other, and form a singleton player $[S_{ij}]$, $S_{ij}=\{i,j\}$. Thus, this game can be considered to have $(n-1)$ players, $N_{ij}=N\setminus\{i,j\}\cup\{[S_{ij}]\}$.
If the player $j$ never participates in the game, then the player $i$ is considered to work individually. Similarly, if the player $i$ is absent
in the game, then the player $j$ is also considered to work individually.
In this way, the interaction between players $i$ and $j$, $I(i,j)$, is defined as the contribution of $[S_{ij}]$, when players $i$ and $j$ cooperate with each other \emph{w.r.t.} the sum of $\phi(i|N\setminus\{j\})$ and $\phi(j|N\setminus\{i\})$, when they work individually~\cite{zhang2020interpreting}, as follows.
\begin{equation}
\begin{aligned}
\label{eqn:Interaction}
I(i,j)&=\phi([S_{ij}]|N_{ij})-\left[\phi(i|N\setminus\{j\})+\phi(j|N\setminus\{i\})\right]\\
&=\sum\nolimits_{T\subseteq N\setminus
\{i,j\}}\frac{(n-t-2)!t!}{(n-1)!}\Delta_{ij}v(T),
\end{aligned}
\end{equation}
where $\Delta_{ij}v(S)\overset{\textrm{def}}{=} v(S\cup\{i,j\}) - v(S\cup \{i\}) - v(S\cup \{j\}) + v(S)$.
Eqn.~\eqref{eqn:Interaction} is essentially equivalent to Eqn.~\eqref{interaction index}, when we consider $S$ in Eqn.~\eqref{interaction index} as $S_{ij}$.
If $I(i,j) > 0$, then players $i$ and $j$ cooperate with each other for a higher contribution, \emph{i.e.} the interaction is positive. If $I(i,j) < 0$, then the interaction between players $i$ and $j$ leads to a lower contribution, \emph{i.e.} the interaction is negative.

\textbf{Interaction components of different orders:}  In this study, we find that the overall interaction $I(i,j)$ can be decomposed into interaction components of different orders $m$, \emph{i.e.} $I(i,j) =\frac{1}{n-1} \sum_{m=0}^{n-2} I^{(m)}(i,j)$. Here, we use $m$ to measure the scale of contexts. $I^{(m)}(i,j)$ reflects the average interactions between players $i$ and $j$ among all contexts with $m$ players.  For example, in the object classification, each pixel of the input image can be considered as a player.
The union of $S \subseteq N$ and $\{i,j\}$ can be regarded as an inference pattern. For example, we can consider a human face as an inference pattern. Let players $i$ and $j$ represent two eyes on this face, and other $m$ players on the face represent an inference pattern. In this way, the interaction between two eyes depends on the context of the face, which is measured as $I^{(m)}(i,j)$.

In particular, when $m$ is small, $I^{(m)}(i,j)$ measures the interaction from inference patterns consisting of very few pixels without knowing the global structures of the object. Whereas, when $m$ is large, $I^{(m)}(i,j)$ corresponds to the interaction from inference patterns computed with relatively rich contextual information, which usually encode global structures of the object. $ I^{(m)}(i,j)$ is defined as
\begin{equation}
\begin{aligned}
\label{eqn:multi-order_interaction}
I^{(m)}(i,j)&=\phi^{(m)}([S_{ij}]|N_{ij})-\left[\phi^{(m)}(i|N\setminus\{j\})+\phi^{(m)}(j|N\setminus\{i\})\right]\\
&=\mathbb{E}_{S\subseteq N\backslash \{i,j\},s=m} [\Delta_{ij}v(S)],
\end{aligned}
\end{equation}
where $\Delta_{ij}v(S)\overset{\textrm{def}}{=} v(S\cup\{i,j\}) - v(S\cup \{i\}) - v(S\cup \{j\}) + v(S)$, and Eqn.\eqref{eqn:multi-order_interaction} represents the additional contribution brought by $i$ and $j$ in the context $S$ with $m$ players. In addition, we have proven that for $ m\in \{0,\dots,n-1\}, I^{(m)}(i,j)$ has following properties.

$\bullet\;$\emph{Linearity property}: If the game $u$ satisfies $u(S)=w(S)+v(S)$, where $v$ and $w$ are another two games. Then, $\forall i,j\in N$, $I_u^{(m)}(i,j)$ can be decomposed into $I_u^{(m)}(i,j) = I_w^{(m)}(i,j) + I_v^{(m)}(i,j)$, and $I_{c \cdot u}^{(m)}(i,j)=c\cdot I_{u}^{(m)}(i,j)$, where $c$ is a scalar value.

$\bullet\;$\emph{Dummy property}: The dummy player $i\in N$ satisfies $\forall S\subseteq N$, $v(S\cup \{i\}) = v(S)+v(\{i\})$. It means that the player $i$ has no interactions with other players, \emph{i.e.} $\forall j \in N$, $I^{(m)}(i,j)=0$.

$\bullet\;$\emph{Symmetry property}: If players $i,j \in N$ have same interactions with other players {\small$\forall S\subseteq N \backslash \{i,j\}$}, then their contributions have same dependence on different contexts, {\small$\forall k \in N\backslash\{i,j\}$, $ I^{(m)}(i,k) = I^{(m)}(j,k)$}.

$\bullet\;$\emph{Marginal contribution property}: if $\forall i,j\in N, i\neq j$, {\small$\phi^{(m+1)}(i|N)-\phi^{(m)}(i|N)\! =\!\!\!\mathop{\mathbb{E}} \limits_{j\in N\backslash \{i\}} [I^{(m)}(i,j)]$}.

$\bullet\;$\emph{Accumulation property}: The contribution of $i\in N$ with contexts of $m$ players can be decomposed into interactions dependent on less than $m$ players, {\small$\phi^{(m)}(i|N) = \mathop{\mathbb{E}} \limits_{j\in N\backslash\{i\}} [\sum_{k=0}^{m-1} I^{(k)}(i,j)]+ \phi^{(0)}(i|N)$}.

$\bullet\;$\emph{ Recursive property}: {\small$\phi^{(n-1)}(i|N) - \phi^{(0)}(i|N)\!=\!\!\!\mathop{\mathbb{E}}\limits_{j\in N\backslash \{i\}} [\sum \limits_{m=0}^{n-2} I^{(m)}(i,j)]\!=\!I(N\backslash\{i\}, i)\!=\!\!\!\sum \limits_{j\in N\backslash \{i\}} I(i,j)$.}

$\bullet\;$\emph{ Efficiency property}: The overall reward of the game can be decomposed into interactions of different orders, \emph{i.e.} {\small$v(N)-v(\emptyset)\!=\!\sum\limits_{i\in N}\phi^{(0)}(i|N) +\sum\limits_{i\in N}\sum\limits_{j\in N\backslash \{i\}} [\sum\limits_{k=0}^{n-2} \frac{n-1-k}{n(n-1)} I^{(k)}(i,j)]$}.

\textbf{Purified interaction components of different orders:}  Note that $I^{(m)}(i,j)$ includes the average interaction benefit from $0$-order contexts to $(n-2)$-order contexts. For example, given a specific context $\{a,b\}$, $I^{(2)}(i,j)$ includes marginal interaction benefits from coalitions $\{i,j\}$, $\{i,j,a\}$, $\{i,j,b\}$ and $\{i,j,a,b\}$, \emph{i.e.} $I^{(2)}(i,j)$ reflects a \textbf{mixture} of interactions of different orders. In order to purify the definition of interactions, we propose $J^{(m)}(i,j)$ to exclusively represent the interaction benefit from contexts with \textbf{exactly} $m$ players. Specifically, $R_T(i,j)$ quantifies interaction benefits from the inference pattern of $T\cup\{i,j\}$, which can be computed as follows.
\begin{equation}
\label{eqn:def_RT}
R_T(i,j)=\sum_{T'\subseteq T}(-1)^{t'-t}\Delta_{ij}v(S)
\end{equation}
Thus, the interaction component $I^{(m)}(i,j)$ can be represented as follows.
\begin{equation}
\label{eqn:R_T}
I^{(m)}(i,j) = \mathbb{E}_{S\subseteq N\backslash \{i,j\}, s = m}[\sum\nolimits_{T\subseteq S} R_T(i,j)].
\end{equation}
Similar to Eqn.(\ref{eqn:R_T}), we define $J^{(m)}(i,j)$ to exclusively measure interactions of the inference pattern with players $i$, $j$ and other $m$ players as
\begin{equation}
\label{eqn:Jm}
J^{(m)}(i,j) = \mathbb{E}_{T\subseteq N\backslash \{i,j\}, t=m}[R_T(i,j)].
\end{equation}
Hence, $J^{(m)}(i,j)$ can be computed recursively as follows.
\begin{equation}
\label{eqn:IJrelation}
I^{(m)}(i,j) =\sum\nolimits_{0 \leq p\leq m}\binom{p}{m}\cdot J^{(p)}(i,j) \Rightarrow J^{(m)}(i,j)=I^{(m)}(i,j)-\sum\nolimits_{p=0}^{m-1}\binom{p}{m}\cdot J^{(p)}(i,j)
\end{equation}

\bibliography{main}
\bibliographystyle{plainnat}
\end{document}